\documentclass[letterpaper, 10 pt, conference]{ieeeconf}  % Comment this line out if you need a4paper

\IEEEoverridecommandlockouts                              % This command is only needed if 
\usepackage{graphics} % for pdf, bitmapped graphics files
\usepackage{epsfig} % for postscript graphics files
\usepackage{mathptmx} % assumes new font selection scheme installed
\usepackage{times} % assumes new font selection scheme installed

\usepackage{caption}
\usepackage[utf8]{inputenc} % allow utf-8 input
\usepackage[T1]{fontenc}    % use 8-bit T1 fonts
\usepackage{hyperref}       % hyperlinks
\usepackage{url}            % simple URL typesetting
\usepackage{booktabs}       % professional-quality tables
\usepackage{amsfonts}       % blackboard math symbols
\usepackage{nicefrac}       % compact symbols for 1/2, etc.
\usepackage{microtype}      % microtypography
\usepackage{xcolor}         % colors
\usepackage{graphicx}
\usepackage{subcaption}
\usepackage{multirow}
\usepackage{amsmath}  % For advanced math typesetting
\usepackage{bm}       % For bold math symbols
\usepackage{marvosym}

\title{\LARGE \bf
High-Quality 3D Creation from A Single Image Using Subject-Specific Knowledge Prior
}

\author{Nan Huang$^{1}$, Ting Zhang$^{2\dagger}$, Yuhui Yuan$^{3}$, Dong Chen$^{3}$ and Shanghang Zhang$^{1}$\textsuperscript{\Letter}
\thanks{$^{1}$ Nan Huang and Shanghang Zhang are with State Key Laboratory of Multimedia Information Processing, School of Computer Science, Peking University 
$^{2}$Ting Zhang is with Beijing Normal University.
$^{3}$Yuhui Yuan and Dong Chen are with Microsoft Research Asia.}
\thanks{$\dagger$ Project Leader: tingzhang@bnu.edu.cn}
\thanks{\Letter \hspace{0.2em} Corresponding Author: shanghang@pku.edu.cn.}
}

\begin{document}

\maketitle
\thispagestyle{empty}
\pagestyle{empty}

%%%%%%%%%%%%%%%%%%%%%%%%%%%%%%%%%%%%%%%%%%%%%%%%%%%%%%%%%%%%%%%%%%%%%%%%%%%%%%%%
\begin{abstract}

In this paper, we address the critical bottleneck in robotics caused by the scarcity of diverse 3D data by presenting a novel two-stage approach for generating high-quality 3D models from a single image. This method is motivated by the need to efficiently expand 3D asset creation, particularly for robotics datasets, where the variety of object types is currently limited compared to general image datasets. Unlike previous methods that primarily rely on general diffusion priors, which often struggle to align with the reference image, our approach leverages subject-specific prior knowledge. By incorporating subject-specific priors in both geometry and texture, we ensure precise alignment between the generated 3D content and the reference object. Specifically, we introduce a shading mode-aware prior into the NeRF optimization process, enhancing the geometry and refining texture in the coarse outputs to achieve superior quality. Extensive experiments demonstrate that our method significantly outperforms prior approaches.
% Our approach is well-suited for applications such as novel view synthesis, text-to-3D, and image-to-3D, particularly in the robotics field where diverse 3D data is essential.

% In this paper, we present a novel two-stage approach that creates high-quality 3D from a single image by utilizing a subject-specific knowledge prior customized by the reference image. Previous approaches primarily rely on a general diffusion prior, which struggles to yield consistent results with the reference image. Instead, we suggest incorporating subject-specific prior knowledge in both geometry and texture to ensure an accurate alignment between the 3D content and the reference subject. We specifically utilize a shading mode-aware prior into the NeRF optimization process to enhance the geometry and refine the texture of coarse results, thereby achieving superior refinement. Extensive experiments showcase that our method outperforms previous works by a substantial margin. It produces faithful 360-degree reconstructions with impressive visual quality, and can be applied for various applications such as novel-view-synthesis, text-to-3D, and image-to-3D in the field of robotics.

\end{abstract}

%%%%%%%%%%%%%%%%%%%%%%%%%%%%%%%%%%%%%%%%%%%%%%%%%%%%%%%%%%%%%%%%%%%%%%%%%%%%%%%%
\section{INTRODUCTION}

%Automatically generating high-fidelity 3D digital content is a fundamental and enduring challenge in the fields of computer vision and computer graphics, enabling a wide array of applications, including gaming, filmmaking, and virtual/augmented reality experiences. One line of methods aim to extend the success achieved in 2D generation into the realm of 3D data modeling~\cite{jun2023shap, wang2023rodin, gupta20233dgen, nichol2022point}. However, due to the scarcity of diverse and extensive data resources,  existing 3D generative networks primarily model objects within a specific class or struggle to produce photo-realism high visual quality. In this paper, we undertake a more challenging task that aims to create high-fidelity 3D content  for arbitrary objects from a single image, democratizing 3D modeling for a wider user base.

% in the wild. We consider the task of creating 3D structures and textures from a singular viewpoint.
%Recent advances in 3D digital content creation has been advanced a lot. but the task of reconstructing from a single image is still lagging.

%build 3D modeling, leveraging a large set of multi-view images with their corresp
%onding camera poses; however, acquiring such data is challenging. 

The development of robotic models is increasingly constrained by a critical bottleneck: the limited diversity of 3D data available for training. While general image datasets, such as ImageNet\cite{imagenet}, contain thousands of object categories, robotics datasets are far less diverse. For instance, ShapeNet\cite{chang2015shapenetinformationrich3dmodel} and GraspNet\cite{fang2020graspnet}, two commonly used datasets, include only 55 and 88 object categories. This limitation restricts the ability of robots to effectively interact with a wide range of real-world objects. To address this, there is an urgent need to expand the variety of 3D assets in robotics datasets, enabling more robust and versatile policy models.

At the same time, we also aim to leverage image-to-3D techniques to improve robot interactions with real-world objects. For example, when a robot captures an image of an object using its 2D camera, we can convert that image into a 3D model, thereby enhancing the robot’s ability to interact with and manipulate the object. This approach has the potential to significantly improve robotic performance in complex, dynamic environments.

However, this task remains deeply intricate due to its inherently ill-posed nature, as there is a substantial domain gap between a single 2D image and the full 3D spatial and textural characteristics of the object. 
% it represents.
Despite its great challenge, recent works~\cite{Tang_2023_ICCV, qian2023magic123, melas2023realfusion,xu2023neurallift} have achieved notable results in image-to-3D creation.
They draw inspiration from text-to-3D methods~\cite{poole2022dreamfusion, wang2023prolificdreamer} which utilize score distillation sampling (SDS) loss to optimize a neural radiance field (NeRF). 
% This SDS loss serves as a guiding principle for novel view synthesis, emulating the way humans utilize priors to infer 3D shape and texture from a single image.
While producing promising results,
they still exhibit noticeable inconsistencies when compared with the reference view, either in context of the geometry (\textit{i.e.}, multi-face issue) or from the texture perspective (novel views often lack fidelity, presenting smooth and disruptive particulars).

%suffer from the multi-face problem or the visual quality in novel views lacks fidelity, exhibiting clear inconsistencies with the subject given by the reference image.
% the implicit scene modeling of NeRF

%leveraging the implicit 3D prior inherent in text-to-image diffusion models, mimicking the human knowledge prior about the world, to optimize neural radiance fields. 
%SDS loss. Their success relies on the general diffusion prior.
%novel view is not reference image guided.

%However,  Previous methods fall short struggle for the multi-face problem and smooth textures due to such general prior. In fact, for 

% When crafting intricate 3D models of a specific object, a professional artist need not be equipped with an exhaustive understanding of all the objects in the world. Rather, the essential requirement lies in possessing specialized knowledge pertaining to that specific subject at hand. This targeted expertise shall be more instrumental in the meticulous creation of 3D assets. 

Despite incorporating several adjustments such as CLIP loss and textual inversion considered in~\cite{Tang_2023_ICCV, qian2023magic123}, 
the supervision of novel views still relies on a general text-to-image model~\cite{ramesh2022hierarchical, rombach2021highresolution, saharia2022photorealistic,balaji2022ediffi,halgren2004glide}, which has challenge in producing consistent results based solely on text, given the inherent difficulty in crafting textual descriptions to cover every detail. Therefore, these approaches struggle to maintain consistency and fall short in reconstructing high-fidelity 3D objects.
This realization has inspired us to explore strategies for cultivating subject-specific knowledge prior, aiming to achieve high-quality image-to-3D.

We employ a two-stage framework, which is a widely recognized way for achieving high-quality image-to-3D. We introduce a subject-specific diffusion prior, customized by the reference image, to fully utilize the geometric and appearance information embedded within the reference image. Building on this prior, we propose shading mode-aware guidance in the NeRF optimization process, and texture enhancement in the refinement stage. We name our approach \emph{Customize-It-3D}, reflecting the utilization of the proposed customized prior. 
% Figure~\ref{fig:teaser} illustrates several examples.

% We propose \emph{Customize-It-3D}, a novel approach that fully unleashes the geometry/appearance information provided by the reference image to promote customized knowledge prior for 3D generation. \emph{Customize-It-3D}, like the previous methods mentioned above, adopts a two-stage framework in a coarse-to-fine manner. However, it is guided with subject-specific knowledge prior.

In the first stage, we optimize a neural radiance field to learn an implicit volume representation. The reference view is directly supervised through rendering loss, specifically pixel differences. For novel views, unlike conventional approaches employing general T2I models for SDS loss, we use a subject-specific T2I model.
%, personalized through a single image.
There has been remarkable success in subject-driven image generation~\cite{ruiz2023dreambooth, kumari2023multi}.
They typically demand at least 3 to 5 images capturing the same subject with varying contexts in order to model the key visual features of the subject.
However, our scenario presents a challenge as we possess only a solitary image, which may misguide the model to a trivial solution (mode collapse).
To surmount this limitation, we propose to exploit multiple modalities extracted from the reference image, namely depth map, mask image, and normal map. %generated by off-the-shelf models~\cite{}.
More importantly, by adjusting the model to incorporate the subject's geometric information, we can more precisely guide the generation of NeRF, especially by considering the shading mode in our supervision.
%This not only mitigates the issue arising from the limited number of reference images but also provides an additional unique advantage that adapts the model to be more aware of the subject's geometry information, thereby facilitating NeRF generation with more fine-grained supervision.
% to personalize the T2I model. This has an additional benefit to adapt the model to NeRF generation needs.
% what normal is and has better adaptation when the output of the NeRF is selected as normal, inject geometry information??

% We introduce a two-stage method that all supervised loss is conditioned on the given reference image

%All these methods do not take
%into account the unique characteristics of object, and utilize fixed strategy for different cases

%In the first stage, we leverage a personalized dreambooth diffusion prior to guide the nerf optimization.
%There has been remarkable success in 
%Unlike the original dreambooth requires 3-5 images about the object. Yet we only have one image which would mislead the model. what is the challenge?
%Therefore, we leverage the other modalities of the ref image to do dreambooth, hopeful to get the better geometry as well as the texture. We add pixel loss at the reference view and subject-specified SDS loss for novel views.

While NeRF is renowned for effectively learning complex geometries, their proficiency in generating intricate texture details is hindered by significant memory demands.
Therefore, our objective in the second stage is to enhance visual realism by transforming the coarse NeRF model into point clouds.
Conversion to point clouds allows us to have the desirable groundtruth textures on the reference view through projection. 
Yet we observe that constructing point clouds from depth images as in~\cite{Tang_2023_ICCV} tends to introduce geometric inaccuracies 
%stemming from misalignment between depth images of different viewpoints. 
since NeRF does not store any 3D geometry explicitly
(only the density field).
In light of this, our strategy entails an initial step of mesh generation, followed by the established mesh regularization techniques to refine the shape, before sampling points on the mesh surface.
In addition, we harness the potential of the aforementioned subject-specific diffusion model, to effectively elevate the texture realism, before extending the RGB rendering from the coarse NeRF to point clouds. This augmentation is shown to yield superior texture quality in the final outcomes.

To evaluate our method, we collect a benchmark including 100 images, and compare with baselines on RealFusion~\cite{melas2023realfusion}, and our 
% Customize100 
dataset.
% Extensive experiments demonstrate our method achieves significant improvement in visual quality in context of accurate geometry and realistic details. 
%Furthermore, our approach supports a range of appliations such as text-to-3D creation and text-guided 3D editing.
In summary, our contributions are:
\begin{itemize}
% [noitemsep,topsep=0pt]
\item We propose \emph{Customize-It-3D}, an innovative framework designed to address the critical bottleneck in robotics caused by the limited diversity of 3D data. 
By generating high-quality 3D content from a single image using a subject-specific diffusion prior, this framework enhances the personalization of 3D asset creation, which is essential for expanding object variety in robotics datasets.
\item We introduce a multi-modal DreamBooth model that promotes comprehensive knowledge priors derived from multi-modal images of the subject, prioritizing the faithfulness of the diffusion model to the reference image. This knowledge prior not only facilitates shading mode-aware NeRF optimization but also enhances the texture for better refinement.
\item The resulting framework demonstrates state-of-the-art performance in 3D reconstruction using a diverse range of in-the-wild images and images from existing datasets.
% achieving significant improvement in visual quality with realistic details and accurate geometry.
 %and text-guided 3D editing, extending its utility beyond image-to-3D task.
\end{itemize}

\section{Related work}
%We mainly review related works that requires prior in 3D modeling.

\noindent \textbf{Muti-view 3D reconstruction.}
Early works~\cite{agarwal2011building,schonberger2016pixelwise,furukawa2015multi,schonberger2016structure} traditionally require multiple input images to deduce geometry by establishing correspondence within the overlapping regions.
The advent of NeRF~\cite{mildenhall2020nerf, lombardi2019neural,mueller2022instant, chen2022tensorf, chen2023dictionary} have significantly propelled the quality via implicit neural representations.
Subsequent efforts~\cite{jain2021putting,kim2022infonerf,du2023learning, yu2021pixelnerf} have been dedicated to facilitate NeRF optimization from a sparser set of input views. Typically they entail the extraction of per-view features from each input image and aggregate multi-view features for each point. This aggregated information is then decoded to determine density (or Signed Distance) and colors.
Diffusion models have emerged as a recent focal point in 3D generation research. 
Some studies endeavor to directly train 3D diffusion models based on diverse 3D representations, including point clouds~\cite{luo2021diffusion,nichol2022point,zeng2022lion,zhou20213d},
meshes~\cite{gao2022get3d,liu2023meshdiffusion}, neural fields~\cite{anciukevivcius2023renderdiffusion,chen2023single,cheng2023sdfusion,erkocc2023hyperdiffusion,gupta20233dgen, jun2023shap,karnewar2023holofusion, kim2023neuralfield, kontschiederdiffrf, gu2023nerfdiff, ntavelis2023autodecoding,wang2023rodin, zhang20233dshape2vecset}.
Others, such as~\cite{chan2023genvs,watson2022novel,szymanowicz2023viewset,shi2023mvdream} 
conceptualize 3D generation as an image-to-image translation and directly generate coherent multiview images, as exemplified by Zero123~\cite{liu2023zero1to3}.
% It's noteworthy that both approaches require access to multi-view images for training.
% A recurring challenge in this domain lies in the constrained availability of expansive 3D assets datasets. Consequently, most studies have been confined to limited categories of shapes or encounter difficulties in extending their models effectively on general objects.
%and how to
%scale up on large datasets is still an open problem. 

\noindent \textbf{3D generation using knowledge prior.}
% A surge of approaches have been study leveraging large models as knowledge prior to guide the 3D synthesis, 
% circumventing the need for large-scale 3D datasets. One popular task is text-to-3D synthesis, given
% the exceptional breakthroughs in text-to-image generation.
Pioneering works~\cite{mohammad2022clip, seo2023ditto} such as
DreamFields~\cite{jain2022zero}, DreamFusion~\cite{poole2022dreamfusion} and SJC~\cite{wang2023score}
% have demonstrated success of incorporating NeRF-like models with frozen generative systems for optimization-based 3D synthesis.
% % They fed the rendered 2D images to diffusion models or CLIP model for calculating losses to guide the 3D shape optimization.
% It sparked numerous follow-up works improve
% such per-shape optimization scheme
% in the context of 3D representations~\cite{lin2023magic3d,tsalicoglou2023textmesh,chen2023fantasia3d, lee2022understanding}, sampling schedules~\cite{huang2023dreamtime}, loss design~\cite{wang2023prolificdreamer,qiu2023richdreamer}, generation speed~\cite{li2023instant3d} and subject-driven editing~\cite{raj2023dreambooth3d}.
% An alternative avenue of
sparked numerous follow-up works which~\cite{melas2023realfusion, Tang_2023_ICCV,qian2023magic123,liu2023one,wang2023imagedream,wu2023hyperdreamer,long2023wonder3d,liu2023one,sun2023dreamcraft3d} seeks to generate 3D digital content from a single image, offering a more precise and controllable approach. There are two categories along this direction. One is data-driven approach that trains a feed-forward model with extensible 3D supervision. For instance, LRM~\cite{hong2023lrm} adopts a large transformer-based encoder-decoder architecture for learning 3D representations of objects from a single image.
DMV3D~\cite{xu2023dmv3d} trains on large scale multi-view image datasets of highly diverse object.
Another existing endeavors in this area remain rooted in optimization-based techniques, imposing pixel-wise reconstruction losses on the reference view, in addition to  the SDS loss.
For example, SyncDreamer~\cite{liu2023syncdreamer} adopts a synchronized multiview diffusion to model the joint probability distribution of multiview images for multiview consistency.
Magic123~\cite{qian2023magic123} uses 2D and 3D priors simultaneously to generate faithful 3D content from any given
image.
DreamGaussian~\cite{tang2023dreamgaussian} designs a generative 3D Gaussian Splatting model
with companioned mesh extraction and texture refinement in UV space, aiming at accelerating the optimization process.

In comparison, we follow the optimization-based pipeline~\cite{poole2022dreamfusion}, but focus on cultivating subject-specific knowledge prior. While Dreambooth3D~\cite{raj2023dreambooth3d} also leverages customized diffusion prior, it requires multiple subject images and is designed for text-to-3D editing. Instead, our approach tackles a more challenging scenario with only one image and introduces a multi-modal diffusion prior.

%One-2-3-45~\cite{} proposes to directly produce 3D geometry from the generated images from zero123~\cite{}. Although the method achieves high efficiency, its results are of low quality and lack geometric details.
%Magic123~\cite{} advocate for SDS loss of a joint 2D and 3D prior.

\begin{figure*}[t]
    \centering
    \vspace{0em}
    \includegraphics[width=1\textwidth]{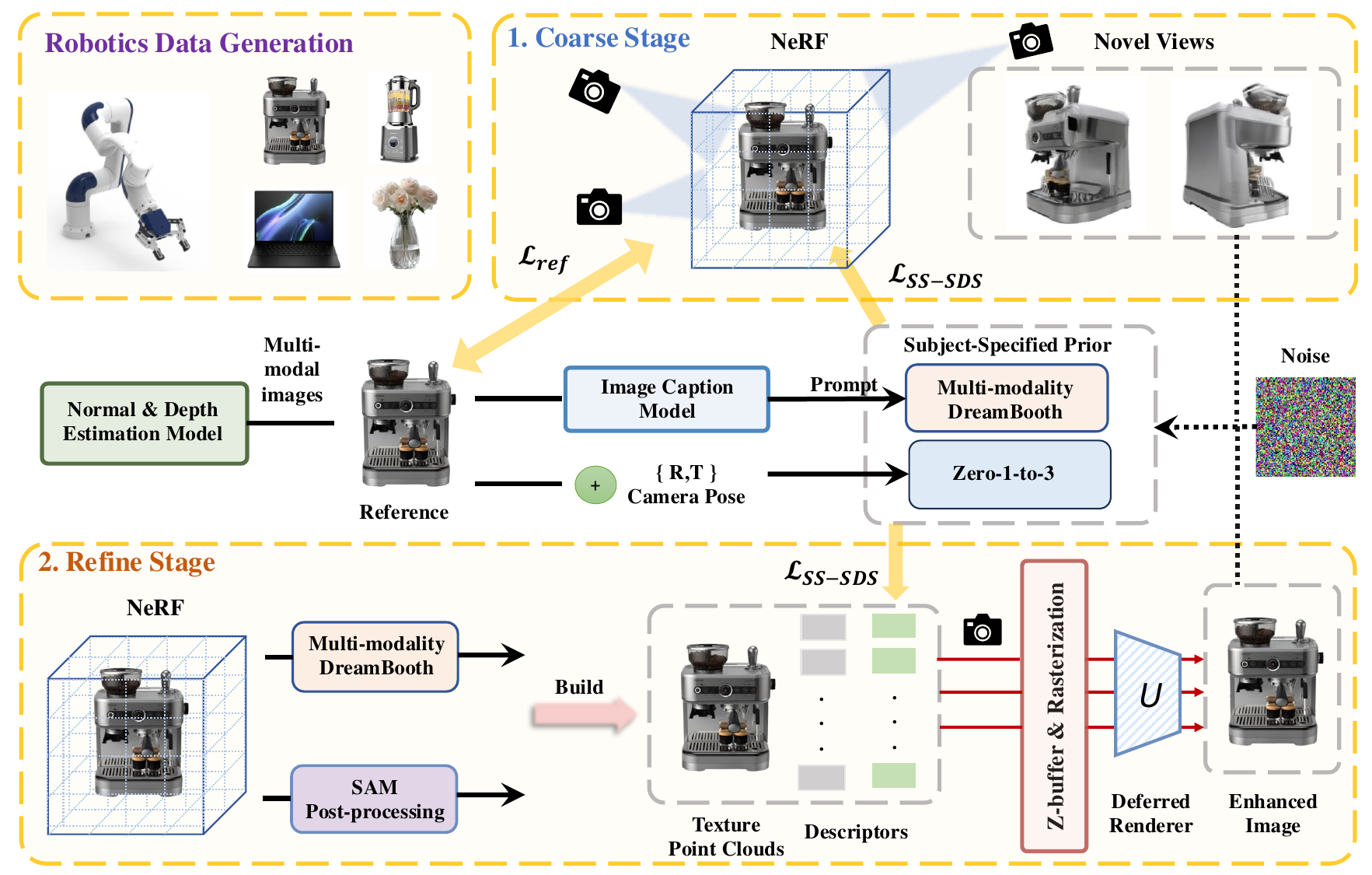}
    \caption{We propose a two-stage framework for 3D creation from a reference
    	image with subject-specific diffusion prior (Sec.~\ref{subsec:joint prior}). At the coarse stage, we  optimize a NeRF for reconstructing the geometry of the
    	reference image in a shading-aware manner(Sec.~\ref{subsec: coarse stage}). We further build point clouds with enhanced texture from the coarse stage, and jointly optimize
    	the texture of invisible points and a learnable deferred renderer to generate realistic and view-consistent textures (Sec.~\ref{subsec:refine stage}).}
    \label{fig:pipeline}
    \vspace{-2em}
\end{figure*}

\section{Method}
\label{sec:methods}

% Generating a personalized 3D object from a single image is highly challenging, because it involves disclosing the identity, geometric shape, and texture from limited and incomplete information. 
In this work, we propose to leverage carefully cultivated subject-specific knowledge prior to effectively constrain the coherency of 3D object to a particular identity to address the need for expanding 3D assets in robotics.
The proposed framework is illustrated in Figure~\ref{fig:pipeline}.
% We will first detail our subject-specific knowledge prior and then introduce methods.
%in the following section.
%for generating high-fidelity and personalized 3D objects from a single image. Specifically, we aim to ensure that the generated 3D object is multi-view consistent, and for this purpose, we utilize joint 2D and 3D priors, with the expectation of leveraging implicit knowledge related to 3D geometry and multi-views within the diffusion model. Additionally, we aim to ensure that the 3D object conforms to a specific identity. So in the 2D prior stage, we fine-tune a pretrained diffusion model to achieve this, while in the 3D prior stage, we use a single image as input to obtain more precise supervision tailored to the subject.

%-------------------------------------------------------------------------
\subsection{Subject-Specific Knowledge Prior}
\label{subsec:joint prior}

\noindent
\textbf{Subject-Specific Prior: Multi-modal DreamBooth.}
Current text-to-image diffusion models possess rich semantic knowledge and 2D image knowledge. DreamFusion~\cite{poole2022dreamfusion} is a pioneering work that utilizes the latent knowledge of these diffusion models to guide a 3D representation optimization. However, due to the abundant imagination of 2D diffusion models, it is challenging to precisely control over the generated geometric shapes, textures, and identities using solely text, resulting in discrepancies and thereby causing confusion in guiding 3D reconstructions.
This phenomenon tends to produce smoothed textures and imprecise geometries as a result of averaging the inconsistent outcomes. To address this issue, we undertake a fine-tuning process on a pre-trained text-to-image diffusion model using a single reference image. 
%We find that training the model with only a single image makes the NeRF optimization difficult to converge.
%, while using a set of such images leads to more converged results.
To mitigate the risks of overfitting and mode collapse, we choose to leverage multi-modal images in our approach.

%Therefore, we fine-tune a pre-trained text-to-image diffusion model to obtain a Muti-modal reference images model, enabling our 2D prior to acquire the ability to simulate the appearance of subjects in the given reference image and synthesize their novel view in specified text prompts, thus obtaining a 3D object conforming to the specific subject.

Specifically, we first utilize a pre-trained monocular depth estimator~\cite{Ranftl2022} and a single-view normal estimator~\cite{eftekhar2021omnidata}~\cite{kar20223d} to obtain the depth and normal maps of the reference image. Furthermore, we segment the last channel of the input image to obtain a mask map. Then, we use this image set and the reference image together to partially fine-tune Stable Diffusion~\cite{rombach2021highresolution} with DreamBooth~\cite{ruiz2023dreambooth}, so that it learns to bind a unique identifier with the specific subject embedded in the output domain of the model.  We also find that a fully trained DreamBooth tends to overfit the subject viewpoints in input images, leading not only to a severe Janus problem but also reducing diversity, as noted in previous work~\cite{raj2023dreambooth3d}. Therefore, we only use the partially fine-tuned results. Unlike DreamBooth~\cite{ruiz2023dreambooth}, we fine-tune not only the UNet but also the text encoder, which leads to better results. 

Additionally, we utilize a prior preservation loss, which enables the supervision of model training during fine-tuning with self-generated images $\mathbf{x}_{o}$ from the original diffusion model.
% $\hat{x_\theta}$.
% using prompt condition vector $c_{pr}$ . 
%This approach prevents language drift and encourages diversity. Thus, giving a prompt $y$ and text encoder $\varepsilon$ , we can get a conditional vector $c = \varepsilon (y) $, which we use to generate an image with a random noise $\epsilon \sim \mathcal{N} (\mathbf{0,I} ) $ . 
% Then our loss to finetune a diffusion model $\bm{\epsilon}_{\phi }$ is :
The loss is,
\begin{equation}
    \begin{split}
        \mathbb{E} _{x,c,\epsilon ,\epsilon',t}[w_t||\bm{\epsilon}_{\phi }{ (\alpha _t\mathbf{x} + \sigma_t \epsilon ,c)-\mathbf{x}||_2^2
    } + \\
    \lambda w_{t'}||\bm{\epsilon}_{\phi } (\alpha _{t'}\mathbf{x}_{o} + \sigma_{t'} \epsilon' ,c_{o})-\mathbf{x}_{o}||_2^2 ],
      \label{eq:dreambooth}
    \end{split}
\end{equation}
where $\mathbf{x}$ is the reference image and $\alpha _t,\sigma _t,w_t $ control the noise schedule, $c_{o}$ is the prompt to generate $x_{ori}$ 
and c is the prompt with a unique identifier for reference image.
% The second term is prior-preservation loss with weight $\lambda$. 

In terms of prompt processing, we need an identifier with a weak prior in both the language model and the diffusion model. Therefore, we follow previous work~\cite{von-platen-etal-2022-diffusers} and use the ``sks'' identifier for stable diffusion~\cite{rombach2021highresolution}. 
Differently, since we are dealing with a set of images with different modalities,  it is imperative not to employ an identical prompt for all images, as this practice can introduce confusion to the model and lead it to erroneously emphasize common distinctive features. 
Our primary objective is not to model the shared aspects across all modalities but rather to equip the model with the capability to discern and accommodate the distinctions arising from different modalities.
Thus, we incorporate additional instructions,
%We also use a specific subject-oriented coarse category name in the prompt. 
and the prompt $c$ for fine-tuning the diffusion model takes the form of ``a depth map / normal map / foreground mask / rgb photo of sks [class name]''.
The choice of the specific instruction is made in accordance with the given modality during training.
In this way, the model is encouraged to perceive the subject from different perspectives.
% , enriching its subject-specific prior. 
%We show in the experiments our Multi-Modal DreamBooth is beneficial 

% \noindent
% \textbf{Subject-Specific 3D Prior: Viewpoint and Image-Conditioned Diffusion Model.} 
% Presently, large-scale pre-trained diffusion models are primarily trained on 2D images, lacking substantial 3D-specific knowledge. Notably, the recent Zero-1-to-3 model~\cite{liu2023zero1to3} stands out as it utilizes fine-tuning on a synthetic 3D dataset~\cite{objaverseXL} to develop a viewpoint-conditioned diffusion model. This model takes a reference image along with related external camera parameters as inputs, enabling the generation of novel views of the same subject based on the reference image. Therefore, it serves as our preferred strategy for establishing a subject-specific 3D prior, contributing to improving 3D consistency.

% thereby learning a generic mechanism for controlling camera viewpoints. This provides us with a 3D prior that contributes to enhancing 3D consistency.
%Moreover, relying solely on a text-guided diffusion model is insufficient for achieving personalization. Thus, we also require reference image guidance.
%Recently, Zero-1-to-3~\cite{liu2023zero1to3} 

%-------------------------------------------------------------------------
\subsection{Coarse Stage: Image to 3D Reconstruction}
\label{subsec: coarse stage}

In the coarse stage, we aim to obtain the approximate geometric structure of the 3D object and the rough texture necessary for constructing a textured point cloud. In this stage, our scene model is a neural field based~\cite{mildenhall2020nerf} on Instant-NGP~\cite{mueller2022instant}. This choice is made because it can handle complex topological changes in a smooth and continuous manner while also enabling the reconstruction of 3D objects in a relatively fast and computationally efficient manner.

\begin{figure}[t]
    \centering
    \vspace{0em}
    \includegraphics[width=0.5\textwidth]{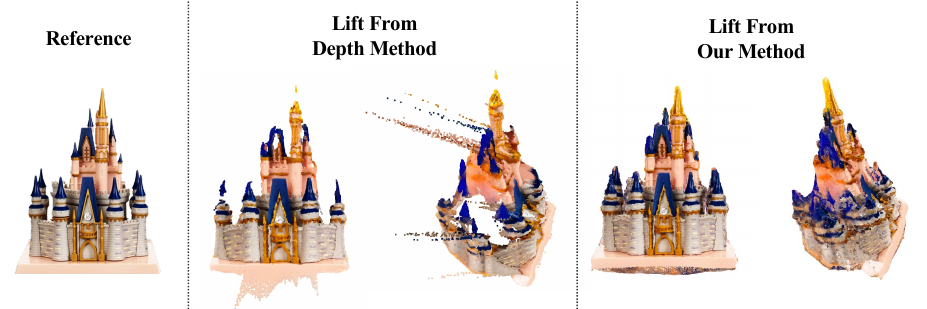}
     \vspace{-1.5em}
    \caption{An example of a Lego castle. Point cloud building results 
    % of point cloud building methods 
    from (1) depth images~\cite{Tang_2023_ICCV} and (2) our method.}
    \label{fig:cloudbuilding}
    \vspace{-2em}
\end{figure}

\noindent
\textbf{Subject-specific knowledge prior for novel views.} 
%Using only the reconstruction loss under the reference view is insufficient, as this may lead to implausible results. Therefore, in the novel view, 
%We follow prior works ~\cite{Xu_2022_neuralLift}~\cite{poole2022dreamfusion} and apply the SDS loss. Different from them, we utilize the aforementioned Subject-Specific Knowledge Prior to guide the optimization of NeRF. 
We optimize the neural
radiance field 
% such that its multi-view renderings look like high-quality samples 
using the subject-specific diffusion model.
Specifically, let 
the rendered image $\mathbf{I} = \mathcal{G}_{\theta }(\mathbf{v} )$ at the viewpoint $\mathbf{v}$, where $\mathcal{G}$ is the differentiable rendering function for NeRF optimization parameterized by $\theta$. We employ the multi-modal DreamBooth model as a 2D prior and follow~\cite{poole2022dreamfusion} ~\cite{lin2023magic3d} to use the score distillation (SDS) loss, which assigns a ``score'' to the rendered image, guiding the optimization of the 3D model's parameters $\theta$ towards the direction of higher density regions.

% \begin{figure}[t]
%     \centering
%     \includegraphics[width=0.9\textwidth]{figure/shading_mode.pdf}
%     \caption{The proposed shading mode-aware guidance largely enhances the 3D geometry in the coarse stage.}
%     \label{fig:shadingmode}
%     \vspace{-1.5em}
% \end{figure}

Differently, we go beyond this and propose modifying the text prompt based on the specific NeRF shading mode. This modification is aimed at fully harnessing the capabilities of our multi-modal diffusion model to offer precise guidance tailored to different NeRF shading modes, which is facilitated by the inherent multi-modal awareness of our finetuned diffusion model.
To elaborate,
given the text prompt $y$ that is generated from an image captioning model~\cite{Li2023BLIP2BL} from the reference image, we adapt the text prompt to 
``sks normal map of $y$" when the NeRF rendering employs ``normal" shading mode
and ``sks rgb photo of $y$" when the NeRF rendering utilizes ``albedo" shading mode.
By enforcing consistency between the rendered image aligned with the modified text prompt, the multi-modal diffusion model is able to provide more accurate guidance and thus enhance the quality of the 3D reconstruction. Formally, we use 2D SDS loss as:
%and use this to perform SDS on the latent space of our multi-modal diffusion model $\bm{\epsilon}_{\phi }$. Let 
%We employ this Muti-modality reference image DreamBooth as a 2D prior and follow prior method~\cite{poole2022dreamfusion} ~\cite{lin2023magic3d} to use the score distillation (SDS) loss.
%Roughly speaking, given a text prompt $y$, we aim to obtain a 3D representation rendering that is as close as possible to the credible generation samples of the T2I model. The SDS loss can 
 %to guide the optimization process of 3D model:
\begin{equation}
	\nabla_{\theta}\mathcal{L}_{SS2D} = \mathbb{E}_{t,\bm{\epsilon} }[w(t)(\bm{\epsilon} _{\phi }(\mathbf{z_t};y_{m},t)- \bm{\epsilon} )\frac{\partial \mathbf{z} }{\partial \mathbf{I}_{m}}  \frac{\partial \mathbf{I}_{m}}{\partial \theta } ] ,
	\label{eq:2d prior}
\end{equation}
where $\mathbf{I}_m = \mathcal{G}_{\theta }(\mathbf{v}, m )$ and $m$ denotes the shading mode and can take on values of either ``normal" or ``albedo".
We use the personalized text prompt $y_m$ to encode NeRF's novel view rendering $\mathbf{I}_m$ to the noisy latent $z_t$ by adding a random Gaussian noise $\bm{\epsilon} $ of a timestep $t$. 
%And $\phi$ is the parameters of our frozen multi-modal DreamBooth model. 

In addition, we leverage a 3D prior for novel view guidance, thanks to the recent Zero-1-to-3 model~\cite{liu2023zero1to3}.
%that utilizes fine-tuning on a synthetic 3D dataset~\cite{objaverseXL} to develop a viewpoint-conditioned diffusion model. 
The 3D loss can be formulated as follows:
\begin{equation}
 \nabla_{\theta}\mathcal{L}_{SS3D} = \mathbb{E} _{t,\epsilon } \left [ w(t)(\epsilon _{\phi }(\mathbf{z_t;x},t,R,T)- \epsilon ) \frac{\partial \mathbf{I}}{\partial \theta }  \right ] ,
\label{eq:3d prior}
\end{equation}
where $R,T$ represent the camera's rotation and translation, and $\mathbf{x}$ stands for the input reference image. 
% Here, we do not utilize text as guidance but instead use image-based guidance, specifically the relative camera transformations between viewpoints.
%The details of this process are presented in \ref{subsec:joint prior} and can be formulated as:
The overall supervision is:
\begin{equation}
\nabla_{\theta}\mathcal{L}_{SS-SDS} = \lambda _ {SS2D}  \nabla_{\theta}\mathcal{L}_{SS2D} + \lambda _ {SS3D}  \nabla_{\theta}\mathcal{L}_{SS3D},
  \label{eq:sds loss}
\end{equation}
where $\lambda _ {SS3D}$ and $\lambda _ {SS3D}$ are their weights respectively. 
%We found that using only the fine-tuned subject-specific 2D prior tends to overfit the reference view, and even with only partial fine-tuning using multi-modality reference images, it is challenging to strike a good balance between preserving identity and maintaining multi-view consistency. Therefore, combining both 2D and 3D priors significantly strengthens the constraint on 3D consistency.
%Moreover, using only 3D priors cannot adequately fit the identity and may reduce diversity, resulting in oversimplified generated 3D objects. Hence, both 2D and 3D priors are essential and complementary, as noted in ~\cite{qian2023magic123}.

\noindent 
\textbf{Groundtruth knowledge for the reference view.} 
Under the reference view $\mathbf{v}_{ref}$ , the rendered image $\mathbf{I} = \mathcal{G}_{\theta }(\mathbf{v} _{ref})$ by NeRF should be consistent with the input image $\mathbf{x}$ to align with our goal of customizing 3D objects. Therefore, we utilize the pixel-wise difference between the NeRF rendering and the input image under the reference view as one of our major losses:
\begin{equation}
	\mathcal{L} _{ref} = ||\mathbf{x}\odot M-\mathcal{G}_{\theta }(\mathbf{v} _{ref}) ||_1,
	\label{eq:ref loss}
\end{equation}
where $\odot$ is Hadamard product. And we follow previous work ~\cite{yariv2020multiview} to apply foreground mask $M$ in order to get extracted object which will ease the geometry reconstruction. 

Using only the RGB per-pixel losses can lead to poor quality geometry issues such as sunken faces, over-flattening, 
%and so on [???], 
due to the inherent shape ambiguity in 3D reconstruction. To mitigate this issue, we employ MiDaS ~\cite{Ranftl2022}, a pretrained monocular depth estimator, to estimate the depth $d_{ref}$ of the reference image. However, since the estimated pseudo depth may not be accurate and there is a scale and source mismatch with the depth $d$ from NeRF, we regularize the negative Pearson correlation between the pseudo depth and the depth from NeRF under the reference viewpoint,
\begin{equation}
\mathcal{L}_{depth} = - \frac{\mathrm{Cov}(d_{ref},d) }{\mathrm{Var}(d_{ref})\mathrm{Var}(d) }  .
  \label{eq:depth loss}
\end{equation}
Here $\mathrm{Cov(.)}$ denotes covariance and $\mathrm{Var(.)}$ measures standard deviation. We employ this function as our depth regularization to make the NeRF output $d$ under the reference view close to the depth prior.

\noindent
\textbf{Overall training.} 
In summary, the loss in the coarse stage consists of $\mathcal{L}_{ref}$, $\mathcal{L}_{SS-SDS}$ and $\mathcal{L}_{depth}$. During training, we follow~\cite{Tang_2023_ICCV}, adopting a progressive training strategy. We start by training only in a partial region of the object's front view, and then gradually expand the training scope. 

%This approach enables more robust training results, {\color{red}{as shown in Figure[???].}}

%-------------------------------------------------------------------------

\begin{figure*}[t]
    \centering
    % \vspace{-1em}
    \includegraphics[width=1\textwidth]{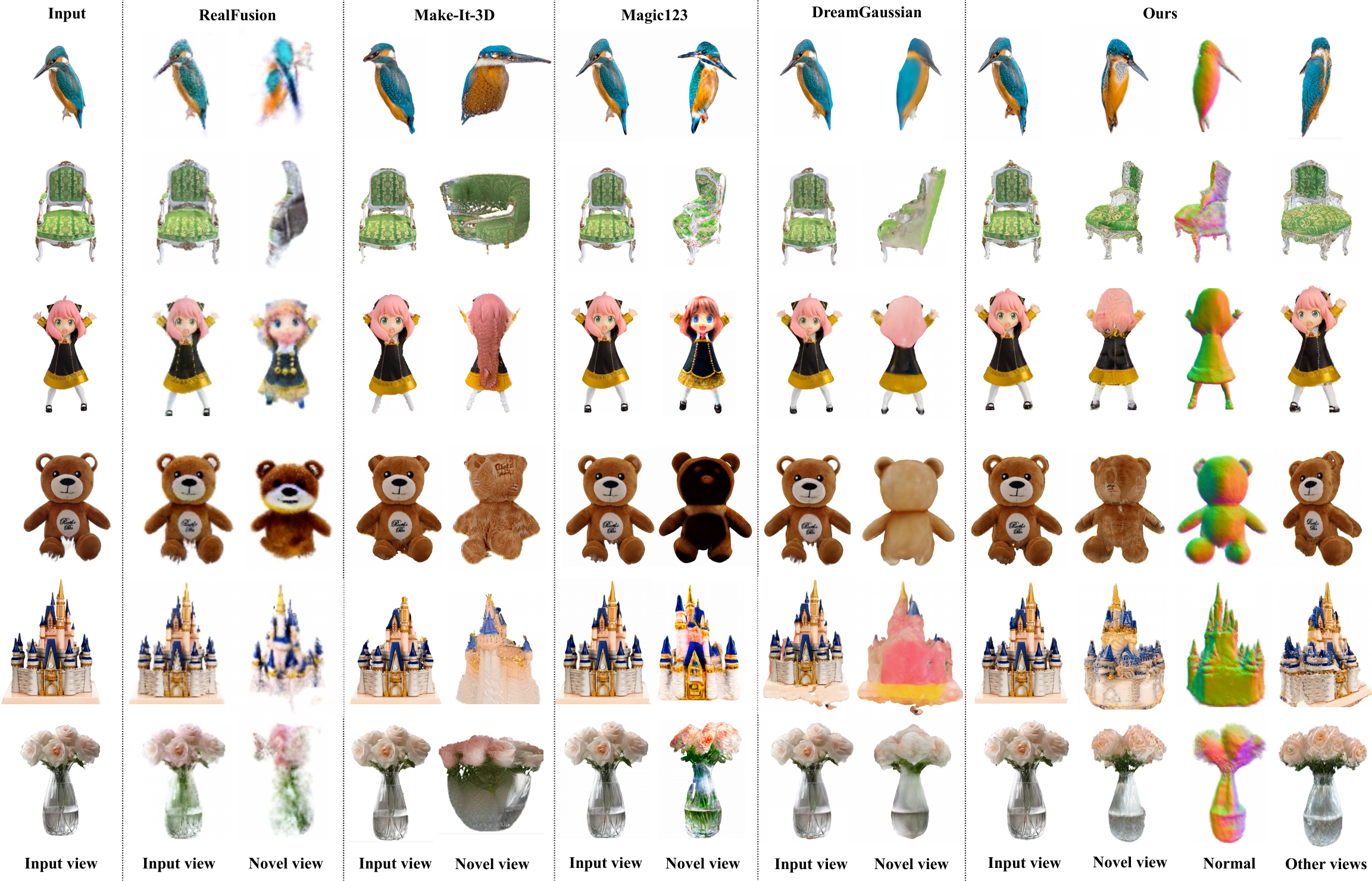}
    \vspace{-1.5em}
    \caption{Qualitative comparison on image-to-3D generation. We compare Customize-It-3D to RealFusion~\cite{melas2023realfusion}, Make-it-3D~\cite{Tang_2023_ICCV} , Magic123~\cite{qian2023magic123} and DreamGaussian~\cite{tang2023dreamgaussian} for
creating 3D objects from a single unposed image (the leftmost column).}
    % \vspace{-1em}
    \label{fig:qualitative}
    \vspace{-0.5em}
\end{figure*}

\begin{figure}[t]
    \centering
    \vspace{-1.3em}
    \includegraphics[width=0.5\textwidth]{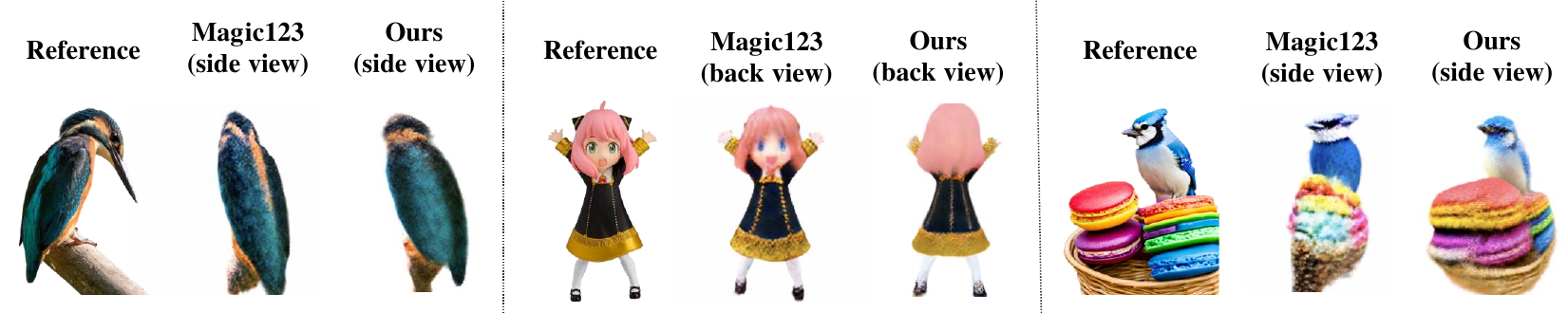}
    \vspace{-1.5em}
    \caption{Comparison with Magic123 in coarse stage. }
    \label{fig:coarse_baseline}
    \vspace{-3em}
\end{figure}

\subsection{Refine Stage: Neural Texture Enhancement}
\label{subsec:refine stage}
In the coarse stage, due to the computationally intensive nature of NeRF optimization, we only perform optimization at low resolutions. Moreover, due to the inherent limitations of NeRF, such as the tendency to produce high-frequency artifacts~\cite{poole2022dreamfusion}, we can only obtain a low-resolution, low-texture-quality 3D model. As a result, we are motivated to explore alternative representations for NeRF, particularly focusing on transforming it into an explicit point cloud. Point clouds have the advantage of allowing direct projection of images, preserving high-quality frontal texture details, and offering opportunities for personalization and enhancement of the projected images, fulfilling the needs of high-quality customized 3D models.
In summary, our goal in the refine stage is to retain the geometric structure of the coarse stage, generate a dense point cloud, and optimize the textures not visible in the reference view $\mathbf{v}_{ref}$, as we directly project the frontal reference image $\mathbf{x}$ onto the point cloud. We elaborate several key designs to facilitate this transformation.

\noindent 
\textbf{Point cloud building.}
Past work has focused on constructing a point cloud from multi-view RGBD images under NeRF rendering~\cite{Tang_2023_ICCV}. However, due to noise in RGBD images and inaccurate depth estimation, the generated point cloud is very noisy 
% and inaccurate
, significantly impairing the  3D geometry in refine stage. To address this, we export the NeRF in coarse stage as a mesh, utilizing mesh regularizations and perform Poisson sampling on the mesh to obtain a dense point cloud $P^{m}$. Figure~\ref{fig:cloudbuilding} shows this difference. We opt to a point cloud because it allows personalized diffusion model-based texture enhancement for each viewpoint, resulting in a higher-quality point cloud that better aligns with our customized 3D object generation needs.

% \begin{figure}
%     \centering
%     \includegraphics[width=1\textwidth]{figure/point_cloud2.pdf}
%     \caption{Illustrating different point cloud building methods from (1) depth images as in~\cite{Tang_2023_ICCV} and (2) our method.}
%     \label{fig:cloudbuilding}
%     \vspace{-1.5em}
% \end{figure}

% \begin{figure}
%     \centering
%     \includegraphics[width=0.5\textwidth]{figure/texture_enhancement2.pdf}
%     \caption{The texture details from coarse rendering are largely enhanced using the multi-modal DreamBooth.}
%     \label{fig:textureenhance}
%     \vspace{-1.5em}
% \end{figure}

\noindent
\textbf{Texture projection.}
However, since NeRF-rendered images from different viewpoints may have overlapping regions, a 3D point may be assigned different colors under different viewpoints~\cite{xie2022high}, leading to texture conflicts. Therefore, 
%inspired by ~\cite{Tang_2023_ICCV}, 
we propose an iterative strategy for constructing texture point cloud $P$ from mesh sampled point cloud $P^{m}$.
Initially, we trust all points  $P^{m}_{ref}$ under reference view $\mathbf{v}_{ref}$ and perform color mapping on $P^{m}_{ref}$, obtaining the front-facing perspective of the new point cloud $P_{ref}$. For novel view projection, we aim to avoid introducing points with color conflicts that overlap with $P_{ref}$.
Hence, we proceed to reproject the recently obtained point cloud $P_{ref}$ onto each novel view $\mathbf{v}_i$ to generate a new mask denoted as $M^{ref\_proj}_{\mathbf{v}_i}$. We then project textures to the points under this novel view only when these points belong to the set difference between the novel view mask $M_{\mathbf{v}_i}$ and the reprojected mask $M^{ref\_proj}_{\mathbf{v}_i}$.
Finally we have a clean point cloud $P = \{p_{ref},p_1,...,p_n\}$ which ensures color mapping is done without introducing conflicts.

% \vspace{-1em}
\noindent
\textbf{Texture and mask enhancement.}
The texture quality obtained in the coarse stage is suboptimal. Therefore, prior to projecting the reference image $\mathbf{x}$ from the reference view and rendered RGB images $\mathcal{X}=\{ \mathbf{x}_{\mathbf{v}_1}, \mathbf{x}_{\mathbf{v}_2},...,\mathbf{x}_{\mathbf{v}_n}\}$ under novel views onto the point cloud geometry, we perform a texture enhancement process on all rendered images within $\mathcal{X}$. Specifically, we add noise to the rendered images and leverage the previously fine-tuned multi-modal personalized diffusion model to denoise into a set of pseudo images $\mathcal{X}^{pseudo}$ that possess higher texture quality as well as multi-view consistency. 
% Figure~\ref{fig:textureenhance} shows that the texture details can be significantly improved by utilizing our multi-modal DreamBooth model.
Simultaneously, due to the issues in the coarse stage of NeRF, the generated masks $\mathcal{M}=\{ M_{\mathbf{v}_1}, M_{\mathbf{v}_2},...,M_{\mathbf{v}_n}\}$ by NeRF are not satisfactory. Fine details (for example long and thin parts) in the object structure may be compromised, leading to defects in point cloud texture and geometry initialization due to mask inadequacies. Therefore, we adopt Segment Anything Model (SAM)~\cite{kirillov2023segany} to generate more refined and accurate masks $\mathcal{M}^{pseudo}$.
%Specifically, for each rendered RGB image, we employ SAM to generate a mask $M_{SAM}$. 
However, due to the insufficient resolution of RGB images and the noise introduced by SAM, we assess the quality of the generated mask by relying on SAM's score. Only when the quality surpasses a certain threshold do we consider using this mask; otherwise, we adhere to utilizing masks generated by NeRF. 
% \vspace{0.5em}
% As shown in Figure \ref{fig:mask_ablation}, our method can better retain the geometry.
% And we illustrate our textured point cloud building in Figure~\ref{fig:pointcloud}.

%We also perform post-mask-processing, wherein we apply certain morphological operations to both SAM-generated and NeRF-generated masks, taking the union of the two masks to complement fine structures lacking in the initial mask. Finally, we have $M_{pseudo} = \hat{M}\cup \hat{M} _{SAM}$, where $\hat{M}$ and $\hat{M} _{sam}$ are the rendering and SAM-generated mask after post-processing under novel view.

%and aim to avoid introducing points with color conflicts that overlap with $p^{m}_{ref}$. We use a reference mask $M^{ref}_{pseudo}$ to perform color mapping on $p^{m}_{ref}$, obtaining the front-facing perspective of the new point cloud $p_{ref}$. For all remaining regions under novel view, we project the just obtained point cloud $p_{ref}$ onto each novel view $v_i$ to create a new mask $m^{re}_i$. We then take the set difference between the novel view mask $m^i_{pseudo}$ and the reprojected mask $m^{re}_i$ as the mask for the remaining points $P_{left} = P^m - p^{m}_{ref}$. Finally we will have a clean point cloud $P = \{p_{ref},p_1,...,p_n\}$ which ensures color mapping is done without introducing conflicts.

\begin{figure}[t]
    \centering
\includegraphics[width=0.5\textwidth]{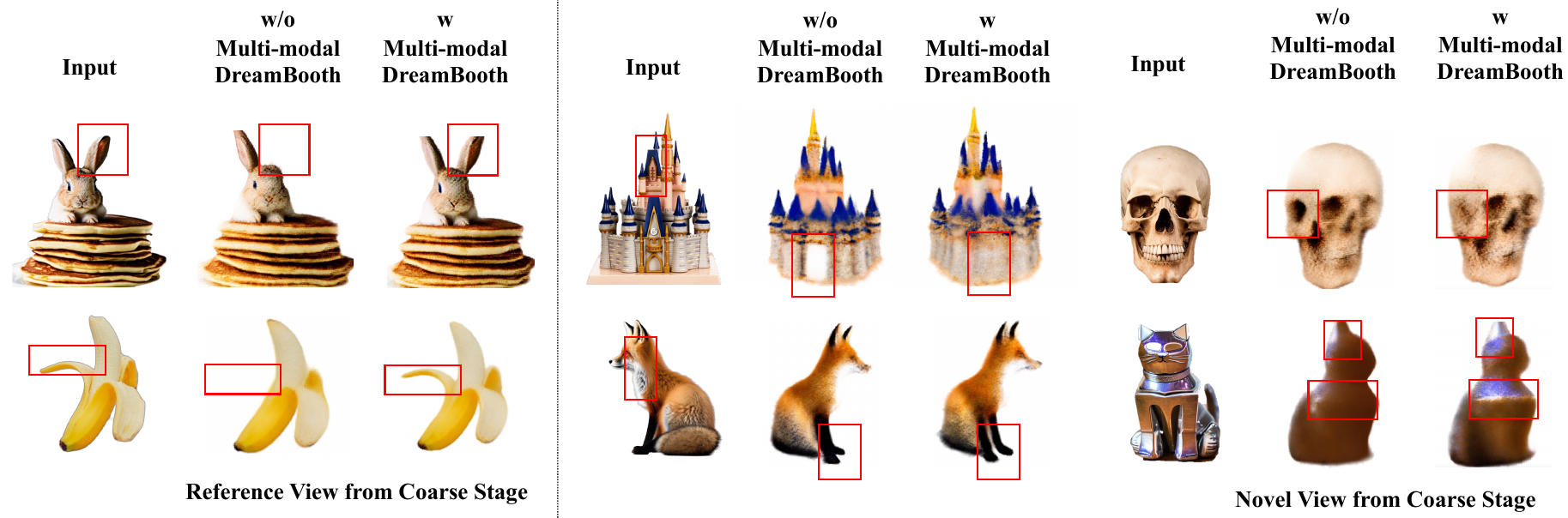}
    \vspace{-1.5em}
    \caption{The effect of multi-modal DreamBooth.
    % showing a consistent improvement. 
    % in terms of both geometry and texture.
    }
    \label{fig:mmdreambooth}
    \vspace{-1em}
\end{figure}

\begin{figure}[t]
    \centering
    % \vspace{-1.5em}
    \includegraphics[width=0.5\textwidth]{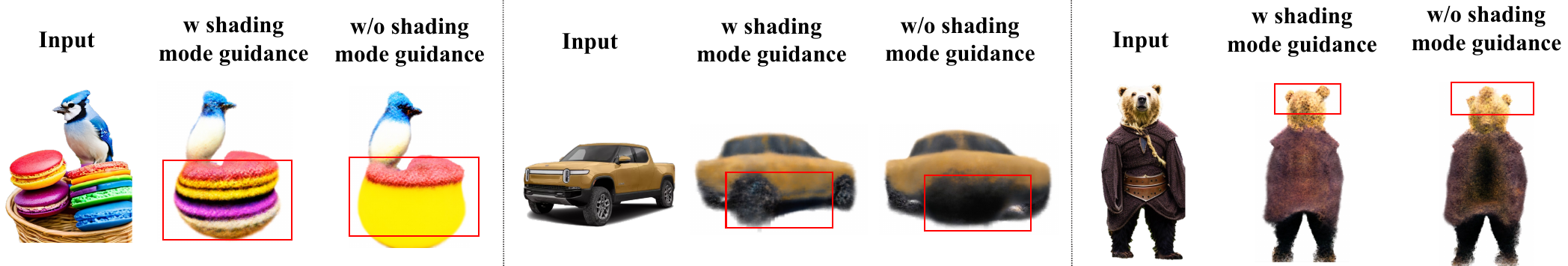}
     \vspace{-1.5em}
    \caption{Our proposed method multi-modal DreamBooth with shading mode-aware guidance shows a clear improvement. All results are obtained under novel view in the coarse stage.}
    \label{fig:singlevsmm}
    \vspace{-2em}
\end{figure}

\begin{figure}[t]
    % \vspace{-1em}
    \centering
    \begin{subfigure}{0.3\linewidth}  % 使用略小于一半栏宽的宽度
        \centering
        \includegraphics[width=\textwidth]{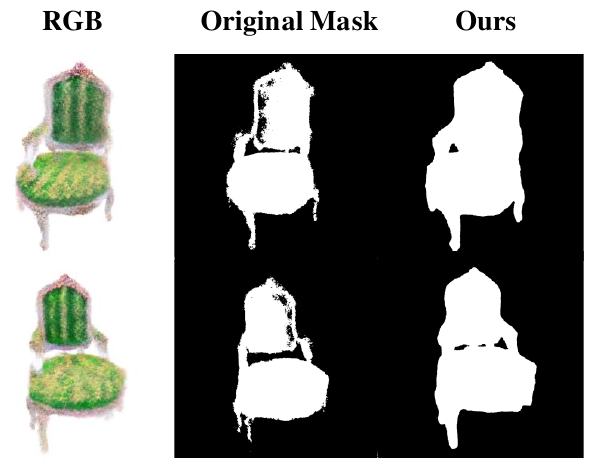}
        \caption{Mask details are better preserved. 
        % by our method.
        }
        \label{fig:mask_ablation}
    \end{subfigure}
    \hspace{0.3em}
    \begin{subfigure}{0.6\linewidth}  % 保持同样的宽度
        \centering
        \includegraphics[width=\textwidth]{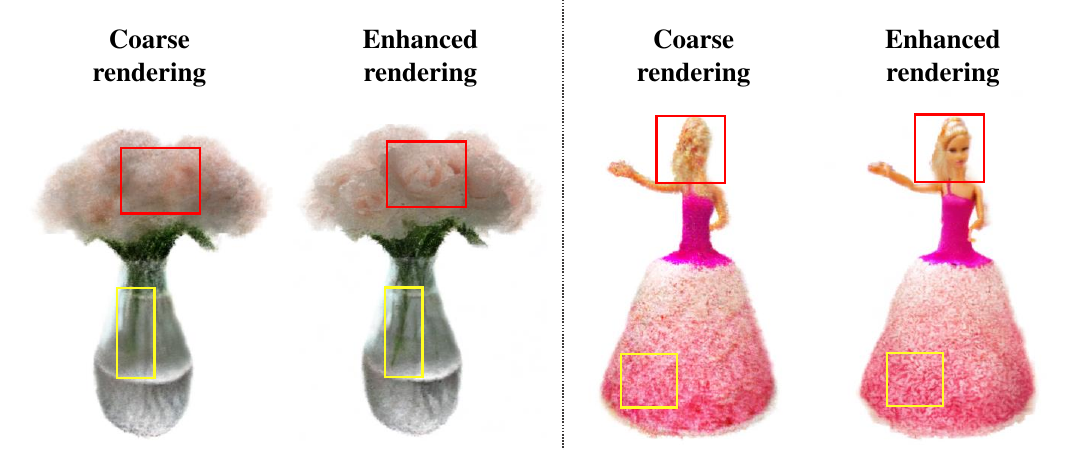}
        \caption{Texture details are enhanced by Multi-modal DreamBooth.}
        \label{fig:textureenhance}
    \end{subfigure}
    \vspace{-0.5em}
    \caption{Results for mask processing and texture enhancement.}
    \vspace{-1em}
    \label{fig:comparative_results}
\end{figure}

\noindent 
\textbf{Deferred point cloud rendering and training.}
Here we adopt a process akin to that described in Make-it-3D~\cite{Tang_2023_ICCV} for point cloud rendering and training, but we incorporate our subject-specific prior.
It is worth to mention that the point cloud building and texture projection is different, which we add an illustration figure in the appendix for better clarity. 

\begin{table}[t]
\centering
% \scriptsize
\resizebox{\linewidth}{!}
{\begin{tabular}{@{\hspace{2pt}}c@{\hspace{2pt}}|@{\hspace{2pt}}c@{\hspace{2pt}}|@{\hspace{2pt}}cccccc}
\toprule
Dataset & Metrics & RealFusion & Make-it-3D& Magic123 & DreamGS  & Ours \\ 
\midrule
\multirow{3}{*}[-0.0ex]{RealFusion} 
& CLIP↑ & 0.73 & 0.83 & 0.82 & 0.77 &  \textbf{0.90} \\
 & PSNR↑ & 16.87 & 20.01 & 19.50 & 18.94 & \textbf{20.50} \\
 & LPIPS↓ & 0.193 & 0.119 & 0.100 & 0.111 &  \textbf{0.094} \\ \midrule
\multirow{3}{*}[0.0ex]{Test-data} 
& CLIP↑ & 0.70 & 0.79 & 0.86 & 0.79 &\textbf{0.88} \\
 & PSNR↑ & 16.41 & 19.31 & 20.10 & 19.80 & \textbf{20.30} \\
 & LPIPS↓ & 0.197 & 0.121 & 0.098 & 0.103 & \textbf{0.096} \\ 
\bottomrule
\end{tabular}}
\vspace{-0.5em}
\caption{Comparison of different methods on RealFusion15 and Test-data. We compute LPIPS and PSNR under the reference view, and CLIP- Score under novel views.}
\label{tab:quantitative}
\vspace{-1.5em}
\end{table}

\begin{table}[h!]
\centering\footnotesize
\begin{tabular}{l|ccc}
\toprule
 & LPIPS$\downarrow$  & PSNR$\uparrow$ & CLIP$\uparrow$ \\ \midrule
% W Single RGB DreamBooth &  0.103  & 20.11  & 80.50$\%$   \\
W/O Multi-modal DreamBooth &  0.107 & 19.88 & 83.70$\%$ \\
\midrule
W Multi-modal DreamBooth (Ours) &\textbf{0.096} & \textbf{20.30}& \textbf{88.08$\%$}\\
\bottomrule
\end{tabular}
\vspace{0em}
\caption{The quantitative ablation study result.}
\label{tab:ablation_quantitative}
\vspace{-2.5em}
\end{table}

\section{Experiments}
\label{sec:experoments}

\subsection{Comparisons with the State of the Arts}
\textbf{Baselines.}
We compare our approach against recent state-of-the-art methods: RealFusion~\cite{melas2023realfusion}, Make-It-3D~\cite{Tang_2023_ICCV}, Magic123 
~\cite{qian2023magic123} and DreamGaussian~\cite{tang2023dreamgaussian}.
% For all methods that utilized Zero123, we employed the Zero123XL checkpoint~\cite{liu2023zero1to3} for fair comparison.
We compare these methods using the Realfusion dataset~\cite{melas2023realfusion} and a customized dataset we created. The Realfusion dataset includes many natural images, while our customized dataset comprises real images and images generated by Stable Diffusion XL~\cite{podell2023sdxl}. 
% We evaluate all the baseline methods using their official code.

\noindent 
\textbf{Qualitative comparison.}
We present a comprehensive collection of qualitative results in Figure~\ref{fig:qualitative}. 
RealFusion often generates flat 3D results with colors and shapes that diverge significantly from the input image.
Make-it-3D exhibits competitive texture quality but suffers from an issue known as long geometry in side views, particularly noticeable in the reconstruction of objects such as chairs.
Magic123 produces visually plausible structures but grapples with a notable issue of multi-face, as it tends to replicate the reference texture in the back view.
The geometry generated by DreamGaussian seems to be highly disordered. Under novel views, the newly generated textures are also of poor quality, lacking in detail.
% In contrast, our approach can reasonably hallucinate the texture details and geometry for novel views that even deviate significantly from the reference image, which greatly improves the fidelity and consistency in creating 3D models.

To further demonstrate the effectiveness of the proposed multi-modal DreamBooth, we present results from the coarse stage without any components in the refine stage in Figure \ref{fig:coarse_baseline} compared to  Magic123\cite{qian2023magic123}.
% the state-of-the-art NeRF-based baseline. 
It can be seen that our results achieve superior results in both geometry and texture, e.g., ours correctly generate the back view of anime girl while Magic123 suffers from multi-face issue.
% Our method accurately generated the back of the cartoon character and also yielded a more seamless texture for the side of the bird.

\noindent 
\textbf{Quantitative comparison.}
We quantitatively compare with the baselines in Table~\ref{tab:quantitative}. We use metrics following~\cite{Tang_2023_ICCV,qian2023magic123} with PSNR, LPIPS~\cite{zhang2018unreasonable}, and CLIP-similarity~\cite{radford2021learning}.
As shown in the table, we achieves best performance. 
% across all the metrics.
% showing that our model is able to generate 3D objects with better 3D consistency.

\section{Conclusions and Discussions}

\noindent 
\textbf{Subject-specific prior: multi-modal DreamBooth.}
We first ablate the effect of using the proposed multi-modal DreamBooth in Figure~\ref{fig:mmdreambooth}. 
%We also compare counterpart that uses single-rgb DreamBooth in the figure. 
We observe a consistent improvement in terms of both texture and geometry. 
Take banana for instance, without multi-modal DreamBooth, the method fails to reconstruct slender structures already visible in the input banana image, which also appears to have unsuccessful reconstruction in Make-it-3D such as the armchair leg shown in Fig \ref{fig:qualitative}. The reason is inconsistency stemmed from SDS loss (novel view) and reconstruction loss (reference view). 
To achieve a balance between these
% two aspects
, Magic123 addresses this with additional 3D prior\cite{liu2023zero1to3}, and 
we alleviates it through multi-modal DreamBooth. This example indeed demonstrate the efficacy of our proposed multi-modal DreamBooth.
Also, we conduct a quantitative study on our test benchmark, validating the efficacy of our approach 
% as shown 
in Table \ref{tab:ablation_quantitative}.
% Employing the suggested multi-modal subject-specific diffusion model prior results in higher-quality 3D content, yielding more compelling visuals with enhanced 3D consistency.

\noindent 
\textbf{Shading-mode-aware guidance.}
We ablate shading-mode guidance and found that with it, our training in the coarse stage converged more easily, showing better performance in both geometry and texture, as illustrated in second column and fourth column of Figure \ref{fig:singlevsmm}. This fully demonstrates that shading-mode guidance effectively utilizes the subject's multi-modal information to provide more robust supervision.

\noindent 
\textbf{Texture and mask enhancement.}
Here we ablate the proposed texture and mask enhancement procedures.
Figure~\ref{fig:textureenhance} shows that the texture details can be significantly improved by utilizing our multi-modal DreamBooth model.
As shown in Figure \ref{fig:mask_ablation}, our method can better retain the geometry.

\section{Acknowledgement}
This work was supported by the National Natural Science Foundation of China (62476011).

\newpage
% \clearpage
\bibliographystyle{splncs04} 
\bibliography{ref}

\begin{thebibliography}{10}
\providecommand{\url}[1]{\texttt{#1}}
\providecommand{\urlprefix}{URL }
\providecommand{\doi}[1]{https://doi.org/#1}

\bibitem{agarwal2011building}
Agarwal, S., Furukawa, Y., Snavely, N., Simon, I., Curless, B., Seitz, S.M., Szeliski, R.: Building rome in a day. Communications of the ACM  \textbf{54}(10),  105--112 (2011)

\bibitem{anciukevivcius2023renderdiffusion}
Anciukevi{\v{c}}ius, T., Xu, Z., Fisher, M., Henderson, P., Bilen, H., Mitra, N.J., Guerrero, P.: Renderdiffusion: Image diffusion for 3d reconstruction, inpainting and generation. In: Proceedings of the IEEE/CVF Conference on Computer Vision and Pattern Recognition. pp. 12608--12618 (2023)

\bibitem{balaji2022ediffi}
Balaji, Y., Nah, S., Huang, X., Vahdat, A., Song, J., Kreis, K., Aittala, M., Aila, T., Laine, S., Catanzaro, B., et~al.: ediffi: Text-to-image diffusion models with an ensemble of expert denoisers. arXiv preprint arXiv:2211.01324  (2022)

\bibitem{chan2023genvs}
Chan, E.R., Nagano, K., Chan, M.A., Bergman, A.W., Park, J.J., Levy, A., Aittala, M., De~Mello, S., Karras, T., Wetzstein, G.: Genvs: Generative novel view synthesis with 3d-aware diffusion models (2023)

\bibitem{chang2015shapenetinformationrich3dmodel}
Chang, A.X., Funkhouser, T., Guibas, L., Hanrahan, P., Huang, Q., Li, Z., Savarese, S., Savva, M., Song, S., Su, H., Xiao, J., Yi, L., Yu, F.: Shapenet: An information-rich 3d model repository (2015), \url{https://arxiv.org/abs/1512.03012}

\bibitem{chen2022tensorf}
Chen, A., Xu, Z., Geiger, A., Yu, J., Su, H.: Tensorf: Tensorial radiance fields. In: European Conference on Computer Vision. pp. 333--350. Springer (2022)

\bibitem{chen2023dictionary}
Chen, A., Xu, Z., Wei, X., Tang, S., Su, H., Geiger, A.: Dictionary fields: Learning a neural basis decomposition. ACM Transactions on Graphics (TOG)  \textbf{42}(4),  1--12 (2023)

\bibitem{chen2023single}
Chen, H., Gu, J., Chen, A., Tian, W., Tu, Z., Liu, L., Su, H.: Single-stage diffusion nerf: A unified approach to 3d generation and reconstruction. arXiv preprint arXiv:2304.06714  (2023)

\bibitem{cheng2023sdfusion}
Cheng, Y.C., Lee, H.Y., Tulyakov, S., Schwing, A.G., Gui, L.Y.: Sdfusion: Multimodal 3d shape completion, reconstruction, and generation. In: Proceedings of the IEEE/CVF Conference on Computer Vision and Pattern Recognition. pp. 4456--4465 (2023)

\bibitem{imagenet}
Deng, J., Dong, W., Socher, R., Li, L.J., Li, K., Fei-Fei, L.: Imagenet: A large-scale hierarchical image database. In: 2009 IEEE Conference on Computer Vision and Pattern Recognition. pp. 248--255 (2009). \doi{10.1109/CVPR.2009.5206848}

\bibitem{du2023learning}
Du, Y., Smith, C., Tewari, A., Sitzmann, V.: Learning to render novel views from wide-baseline stereo pairs. In: Proceedings of the IEEE/CVF Conference on Computer Vision and Pattern Recognition. pp. 4970--4980 (2023)

\bibitem{eftekhar2021omnidata}
Eftekhar, A., Sax, A., Malik, J., Zamir, A.: Omnidata: A scalable pipeline for making multi-task mid-level vision datasets from 3d scans. In: Proceedings of the IEEE/CVF International Conference on Computer Vision. pp. 10786--10796 (2021)

\bibitem{erkocc2023hyperdiffusion}
Erko{\c{c}}, Z., Ma, F., Shan, Q., Nie{\ss}ner, M., Dai, A.: Hyperdiffusion: Generating implicit neural fields with weight-space diffusion. arXiv preprint arXiv:2303.17015  (2023)

\bibitem{fang2020graspnet}
Fang, H.S., Wang, C., Gou, M., Lu, C.: Graspnet-1billion: A large-scale benchmark for general object grasping. In: Proceedings of the IEEE/CVF Conference on Computer Vision and Pattern Recognition. pp. 11444--11453 (2020)

\bibitem{furukawa2015multi}
Furukawa, Y., Hern{\'a}ndez, C., et~al.: Multi-view stereo: A tutorial. Foundations and Trends{\textregistered} in Computer Graphics and Vision  \textbf{9}(1-2),  1--148 (2015)

\bibitem{gao2022get3d}
Gao, J., Shen, T., Wang, Z., Chen, W., Yin, K., Li, D., Litany, O., Gojcic, Z., Fidler, S.: Get3d: A generative model of high quality 3d textured shapes learned from images. Advances In Neural Information Processing Systems  \textbf{35},  31841--31854 (2022)

\bibitem{gu2023nerfdiff}
Gu, J., Trevithick, A., Lin, K.E., Susskind, J.M., Theobalt, C., Liu, L., Ramamoorthi, R.: Nerfdiff: Single-image view synthesis with nerf-guided distillation from 3d-aware diffusion. In: International Conference on Machine Learning. pp. 11808--11826. PMLR (2023)

\bibitem{gupta20233dgen}
Gupta, A., Xiong, W., Nie, Y., Jones, I., O{\u{g}}uz, B.: 3dgen: Triplane latent diffusion for textured mesh generation. arXiv preprint arXiv:2303.05371  (2023)

\bibitem{halgren2004glide}
Halgren, T.A., Murphy, R.B., Friesner, R.A., Beard, H.S., Frye, L.L., Pollard, W.T., Banks, J.L.: Glide: a new approach for rapid, accurate docking and scoring. 2. enrichment factors in database screening. Journal of medicinal chemistry  \textbf{47}(7),  1750--1759 (2004)

\bibitem{hong2023lrm}
Hong, Y., Zhang, K., Gu, J., Bi, S., Zhou, Y., Liu, D., Liu, F., Sunkavalli, K., Bui, T., Tan, H.: Lrm: Large reconstruction model for single image to 3d. arXiv preprint arXiv:2311.04400  (2023)

\bibitem{jain2022zero}
Jain, A., Mildenhall, B., Barron, J.T., Abbeel, P., Poole, B.: Zero-shot text-guided object generation with dream fields. In: Proceedings of the IEEE/CVF Conference on Computer Vision and Pattern Recognition. pp. 867--876 (2022)

\bibitem{jain2021putting}
Jain, A., Tancik, M., Abbeel, P.: Putting nerf on a diet: Semantically consistent few-shot view synthesis. In: Proceedings of the IEEE/CVF International Conference on Computer Vision. pp. 5885--5894 (2021)

\bibitem{jun2023shap}
Jun, H., Nichol, A.: Shap-e: Generating conditional 3d implicit functions. arXiv preprint arXiv:2305.02463  (2023)

\bibitem{kar20223d}
Kar, O.F., Yeo, T., Atanov, A., Zamir, A.: 3d common corruptions and data augmentation. In: Proceedings of the IEEE/CVF Conference on Computer Vision and Pattern Recognition. pp. 18963--18974 (2022)

\bibitem{karnewar2023holofusion}
Karnewar, A., Mitra, N.J., Vedaldi, A., Novotny, D.: Holofusion: Towards photo-realistic 3d generative modeling. In: Proceedings of the IEEE/CVF International Conference on Computer Vision. pp. 22976--22985 (2023)

\bibitem{kim2022infonerf}
Kim, M., Seo, S., Han, B.: Infonerf: Ray entropy minimization for few-shot neural volume rendering. In: Proceedings of the IEEE/CVF Conference on Computer Vision and Pattern Recognition. pp. 12912--12921 (2022)

\bibitem{kim2023neuralfield}
Kim, S.W., Brown, B., Yin, K., Kreis, K., Schwarz, K., Li, D., Rombach, R., Torralba, A., Fidler, S.: Neuralfield-ldm: Scene generation with hierarchical latent diffusion models. In: Proceedings of the IEEE/CVF Conference on Computer Vision and Pattern Recognition. pp. 8496--8506 (2023)

\bibitem{kirillov2023segany}
Kirillov, A., Mintun, E., Ravi, N., Mao, H., Rolland, C., Gustafson, L., Xiao, T., Whitehead, S., Berg, A.C., Lo, W.Y., Doll{\'a}r, P., Girshick, R.: Segment anything. arXiv:2304.02643  (2023)

\bibitem{kontschiederdiffrf}
Kontschieder, P., Nie{\ss}ner, M.: Diffrf: Rendering-guided 3d radiance field diffusion-supplementary document

\bibitem{kumari2023multi}
Kumari, N., Zhang, B., Zhang, R., Shechtman, E., Zhu, J.Y.: Multi-concept customization of text-to-image diffusion. In: Proceedings of the IEEE/CVF Conference on Computer Vision and Pattern Recognition. pp. 1931--1941 (2023)

\bibitem{Li2023BLIP2BL}
Li, J., Li, D., Savarese, S., Hoi, S.C.H.: Blip-2: Bootstrapping language-image pre-training with frozen image encoders and large language models. ArXiv  \textbf{abs/2301.12597} (2023), \url{https://api.semanticscholar.org/CorpusID:256390509}

\bibitem{lin2023magic3d}
Lin, C.H., Gao, J., Tang, L., Takikawa, T., Zeng, X., Huang, X., Kreis, K., Fidler, S., Liu, M.Y., Lin, T.Y.: Magic3d: High-resolution text-to-3d content creation. In: IEEE Conference on Computer Vision and Pattern Recognition ({CVPR}) (2023)

\bibitem{liu2023one}
Liu, M., Shi, R., Chen, L., Zhang, Z., Xu, C., Wei, X., Chen, H., Zeng, C., Gu, J., Su, H.: One-2-3-45++: Fast single image to 3d objects with consistent multi-view generation and 3d diffusion. arXiv preprint arXiv:2311.07885  (2023)

\bibitem{liu2023zero1to3}
Liu, R., Wu, R., Hoorick, B.V., Tokmakov, P., Zakharov, S., Vondrick, C.: Zero-1-to-3: Zero-shot one image to 3d object (2023)

\bibitem{liu2023syncdreamer}
Liu, Y., Lin, C., Zeng, Z., Long, X., Liu, L., Komura, T., Wang, W.: Syncdreamer: Generating multiview-consistent images from a single-view image. arXiv preprint arXiv:2309.03453  (2023)

\bibitem{liu2023meshdiffusion}
Liu, Z., Feng, Y., Black, M.J., Nowrouzezahrai, D., Paull, L., Liu, W.: Meshdiffusion: Score-based generative 3d mesh modeling. arXiv preprint arXiv:2303.08133  (2023)

\bibitem{lombardi2019neural}
Lombardi, S., Simon, T., Saragih, J., Schwartz, G., Lehrmann, A., Sheikh, Y.: Neural volumes: Learning dynamic renderable volumes from images. arXiv preprint arXiv:1906.07751  (2019)

\bibitem{long2023wonder3d}
Long, X., Guo, Y.C., Lin, C., Liu, Y., Dou, Z., Liu, L., Ma, Y., Zhang, S.H., Habermann, M., Theobalt, C., et~al.: Wonder3d: Single image to 3d using cross-domain diffusion. arXiv preprint arXiv:2310.15008  (2023)

\bibitem{luo2021diffusion}
Luo, S., Hu, W.: Diffusion probabilistic models for 3d point cloud generation. In: Proceedings of the IEEE/CVF Conference on Computer Vision and Pattern Recognition. pp. 2837--2845 (2021)

\bibitem{melas2023realfusion}
Melas-Kyriazi, L., Laina, I., Rupprecht, C., Vedaldi, A.: Realfusion: 360deg reconstruction of any object from a single image. In: Proceedings of the IEEE/CVF Conference on Computer Vision and Pattern Recognition. pp. 8446--8455 (2023)

\bibitem{mildenhall2020nerf}
Mildenhall, B., Srinivasan, P.P., Tancik, M., Barron, J.T., Ramamoorthi, R., Ng, R.: Nerf: Representing scenes as neural radiance fields for view synthesis. In: ECCV (2020)

\bibitem{mohammad2022clip}
Mohammad~Khalid, N., Xie, T., Belilovsky, E., Popa, T.: Clip-mesh: Generating textured meshes from text using pretrained image-text models. In: SIGGRAPH Asia 2022 conference papers. pp.~1--8 (2022)

\bibitem{mueller2022instant}
M\"uller, T., Evans, A., Schied, C., Keller, A.: Instant neural graphics primitives with a multiresolution hash encoding. ACM Trans. Graph.  \textbf{41}(4),  102:1--102:15 (Jul 2022). \doi{10.1145/3528223.3530127}, \url{https://doi.org/10.1145/3528223.3530127}

\bibitem{nichol2022point}
Nichol, A., Jun, H., Dhariwal, P., Mishkin, P., Chen, M.: Point-e: A system for generating 3d point clouds from complex prompts. arXiv preprint arXiv:2212.08751  (2022)

\bibitem{ntavelis2023autodecoding}
Ntavelis, E., Siarohin, A., Olszewski, K., Wang, C., Van~Gool, L., Tulyakov, S.: Autodecoding latent 3d diffusion models. arXiv preprint arXiv:2307.05445  (2023)

\bibitem{von-platen-etal-2022-diffusers}
von Platen, P., Patil, S., Lozhkov, A., Cuenca, P., Lambert, N., Rasul, K., Davaadorj, M., Wolf, T.: Diffusers: State-of-the-art diffusion models. \url{https://github.com/huggingface/diffusers} (2022)

\bibitem{podell2023sdxl}
Podell, D., English, Z., Lacey, K., Blattmann, A., Dockhorn, T., Müller, J., Penna, J., Rombach, R.: Sdxl: Improving latent diffusion models for high-resolution image synthesis (2023)

\bibitem{poole2022dreamfusion}
Poole, B., Jain, A., Barron, J.T., Mildenhall, B.: Dreamfusion: Text-to-3d using 2d diffusion. arXiv  (2022)

\bibitem{qian2023magic123}
Qian, G., Mai, J., Hamdi, A., Ren, J., Siarohin, A., Li, B., Lee, H.Y., Skorokhodov, I., Wonka, P., Tulyakov, S., Ghanem, B.: Magic123: One image to high-quality 3d object generation using both 2d and 3d diffusion priors. arXiv preprint arXiv:2306.17843  (2023)

\bibitem{radford2021learning}
Radford, A., Kim, J.W., Hallacy, C., Ramesh, A., Goh, G., Agarwal, S., Sastry, G., Askell, A., Mishkin, P., Clark, J., et~al.: Learning transferable visual models from natural language supervision. In: International conference on machine learning. pp. 8748--8763. PMLR (2021)

\bibitem{raj2023dreambooth3d}
Raj, A., Kaza, S., Poole, B., Niemeyer, M., Mildenhall, B., Ruiz, N., Zada, S., Aberman, K., Rubenstein, M., Barron, J., Li, Y., Jampani, V.: Dreambooth3d: Subject-driven text-to-3d generation. ICCV  (2023)

\bibitem{ramesh2022hierarchical}
Ramesh, A., Dhariwal, P., Nichol, A., Chu, C., Chen, M.: Hierarchical text-conditional image generation with clip latents. arXiv preprint arXiv:2204.06125  \textbf{1}(2), ~3 (2022)

\bibitem{Ranftl2022}
Ranftl, R., Lasinger, K., Hafner, D., Schindler, K., Koltun, V.: Towards robust monocular depth estimation: Mixing datasets for zero-shot cross-dataset transfer. IEEE Transactions on Pattern Analysis and Machine Intelligence  \textbf{44}(3) (2022)

\bibitem{rombach2021highresolution}
Rombach, R., Blattmann, A., Lorenz, D., Esser, P., Ommer, B.: High-resolution image synthesis with latent diffusion models (2021)

\bibitem{ruiz2023dreambooth}
Ruiz, N., Li, Y., Jampani, V., Pritch, Y., Rubinstein, M., Aberman, K.: Dreambooth: Fine tuning text-to-image diffusion models for subject-driven generation. In: Proceedings of the IEEE/CVF Conference on Computer Vision and Pattern Recognition (2023)

\bibitem{saharia2022photorealistic}
Saharia, C., Chan, W., Saxena, S., Li, L., Whang, J., Denton, E.L., Ghasemipour, K., Gontijo~Lopes, R., Karagol~Ayan, B., Salimans, T., et~al.: Photorealistic text-to-image diffusion models with deep language understanding. Advances in Neural Information Processing Systems  \textbf{35},  36479--36494 (2022)

\bibitem{schonberger2016structure}
Schonberger, J.L., Frahm, J.M.: Structure-from-motion revisited. In: Proceedings of the IEEE conference on computer vision and pattern recognition. pp. 4104--4113 (2016)

\bibitem{schonberger2016pixelwise}
Sch{\"o}nberger, J.L., Zheng, E., Frahm, J.M., Pollefeys, M.: Pixelwise view selection for unstructured multi-view stereo. In: Computer Vision--ECCV 2016: 14th European Conference, Amsterdam, The Netherlands, October 11-14, 2016, Proceedings, Part III 14. pp. 501--518. Springer (2016)

\bibitem{seo2023ditto}
Seo, H., Kim, H., Kim, G., Chun, S.Y.: Ditto-nerf: Diffusion-based iterative text to omni-directional 3d model. arXiv preprint arXiv:2304.02827  (2023)

\bibitem{shi2023mvdream}
Shi, Y., Wang, P., Ye, J., Long, M., Li, K., Yang, X.: Mvdream: Multi-view diffusion for 3d generation. arXiv preprint arXiv:2308.16512  (2023)

\bibitem{sun2023dreamcraft3d}
Sun, J., Zhang, B., Shao, R., Wang, L., Liu, W., Xie, Z., Liu, Y.: Dreamcraft3d: Hierarchical 3d generation with bootstrapped diffusion prior. arXiv preprint arXiv:2310.16818  (2023)

\bibitem{szymanowicz2023viewset}
Szymanowicz, S., Rupprecht, C., Vedaldi, A.: Viewset diffusion:(0-) image-conditioned 3d generative models from 2d data. arXiv preprint arXiv:2306.07881  (2023)

\bibitem{tang2023dreamgaussian}
Tang, J., Ren, J., Zhou, H., Liu, Z., Zeng, G.: Dreamgaussian: Generative gaussian splatting for efficient 3d content creation. arXiv preprint arXiv:2309.16653  (2023)

\bibitem{Tang_2023_ICCV}
Tang, J., Wang, T., Zhang, B., Zhang, T., Yi, R., Ma, L., Chen, D.: Make-it-3d: High-fidelity 3d creation from a single image with diffusion prior. In: Proceedings of the IEEE/CVF International Conference on Computer Vision (ICCV). pp. 22819--22829 (October 2023)

\bibitem{wang2023score}
Wang, H., Du, X., Li, J., Yeh, R.A., Shakhnarovich, G.: Score jacobian chaining: Lifting pretrained 2d diffusion models for 3d generation. In: Proceedings of the IEEE/CVF Conference on Computer Vision and Pattern Recognition. pp. 12619--12629 (2023)

\bibitem{wang2023imagedream}
Wang, P., Shi, Y.: Imagedream: Image-prompt multi-view diffusion for 3d generation. arXiv preprint arXiv:2312.02201  (2023)

\bibitem{wang2023rodin}
Wang, T., Zhang, B., Zhang, T., Gu, S., Bao, J., Baltrusaitis, T., Shen, J., Chen, D., Wen, F., Chen, Q., et~al.: Rodin: A generative model for sculpting 3d digital avatars using diffusion. In: Proceedings of the IEEE/CVF Conference on Computer Vision and Pattern Recognition. pp. 4563--4573 (2023)

\bibitem{wang2023prolificdreamer}
Wang, Z., Lu, C., Wang, Y., Bao, F., Li, C., Su, H., Zhu, J.: Prolificdreamer: High-fidelity and diverse text-to-3d generation with variational score distillation. arXiv preprint arXiv:2305.16213  (2023)

\bibitem{watson2022novel}
Watson, D., Chan, W., Martin-Brualla, R., Ho, J., Tagliasacchi, A., Norouzi, M.: Novel view synthesis with diffusion models. arXiv preprint arXiv:2210.04628  (2022)

\bibitem{wu2023hyperdreamer}
Wu, T., Li, Z., Yang, S., Zhang, P., Pan, X., Wang, J., Lin, D., Liu, Z.: Hyperdreamer: Hyper-realistic 3d content generation and editing from a single image. In: SIGGRAPH Asia 2023 Conference Papers. pp. 1--10 (2023)

\bibitem{xie2022high}
Xie, J., Ouyang, H., Piao, J., Lei, C., Chen, Q.: High-fidelity 3d gan inversion by pseudo-multi-view optimization. arXiv preprint arXiv:2211.15662  (2022)

\bibitem{xu2023neurallift}
Xu, D., Jiang, Y., Wang, P., Fan, Z., Wang, Y., Wang, Z.: Neurallift-360: Lifting an in-the-wild 2d photo to a 3d object with 360deg views. In: Proceedings of the IEEE/CVF Conference on Computer Vision and Pattern Recognition. pp. 4479--4489 (2023)

\bibitem{xu2023dmv3d}
Xu, Y., Tan, H., Luan, F., Bi, S., Wang, P., Li, J., Shi, Z., Sunkavalli, K., Wetzstein, G., Xu, Z., et~al.: Dmv3d: Denoising multi-view diffusion using 3d large reconstruction model. arXiv preprint arXiv:2311.09217  (2023)

\bibitem{yariv2020multiview}
Yariv, L., Kasten, Y., Moran, D., Galun, M., Atzmon, M., Ronen, B., Lipman, Y.: Multiview neural surface reconstruction by disentangling geometry and appearance. Advances in Neural Information Processing Systems  \textbf{33} (2020)

\bibitem{yu2021pixelnerf}
Yu, A., Ye, V., Tancik, M., Kanazawa, A.: pixelnerf: Neural radiance fields from one or few images. In: Proceedings of the IEEE/CVF Conference on Computer Vision and Pattern Recognition. pp. 4578--4587 (2021)

\bibitem{zeng2022lion}
Zeng, X., Vahdat, A., Williams, F., Gojcic, Z., Litany, O., Fidler, S., Kreis, K.: Lion: Latent point diffusion models for 3d shape generation. arXiv preprint arXiv:2210.06978  (2022)

\bibitem{zhang20233dshape2vecset}
Zhang, B., Tang, J., Niessner, M., Wonka, P.: 3dshape2vecset: A 3d shape representation for neural fields and generative diffusion models. arXiv preprint arXiv:2301.11445  (2023)

\bibitem{zhang2018unreasonable}
Zhang, R., Isola, P., Efros, A.A., Shechtman, E., Wang, O.: The unreasonable effectiveness of deep features as a perceptual metric. In: Proceedings of the IEEE conference on computer vision and pattern recognition. pp. 586--595 (2018)

\bibitem{zhou20213d}
Zhou, L., Du, Y., Wu, J.: 3d shape generation and completion through point-voxel diffusion. In: Proceedings of the IEEE/CVF International Conference on Computer Vision. pp. 5826--5835 (2021)

\end{thebibliography}

% \end{thebibliography}

\end{document}